\documentclass[manyauthors]{fundam}

\usepackage{url} 
\usepackage[ruled,lined,linesnumbered]{algorithm2e}
\usepackage{graphicx}
\usepackage{amsfonts,amssymb}
\usepackage{wrapfig}
\usepackage{xcolor}
\usepackage{subcaption}
\usepackage{paralist, tabularx}
\usepackage[numbers]{natbib}
\usepackage{scalerel}
\newcommand\doverline[1]{\ThisStyle{%
		\setbox0=\hbox{$\SavedStyle\overline{#1}$}%
		\ht0=\dimexpr\ht0-.15ex\relax
		\overline{\copy0}%
}}

\newcommand{\post}[1]{#1^{\bullet}}
\newcommand{\pre}[1]{{}^{\bullet}#1}

\newcommand{\tu}[1]{\stackrel{#1}{\to}}

\newcommand{\init}{\mathit{i}}
\newcommand{\fin}{\mathit{fin}}
\newcommand{\TS}{\mathord{\textit{TS}}}
\newcommand{\ent}{\mathit{ent}}
\newcommand{\exit}{\mathit{exit}}
\newcommand{\SSP}{\mathit{SSP}}
\newcommand{\ESSP}{\mathit{ESSP}}

\newcommand{\newtext}[1]{\textcolor{black}{{#1}}}

\usepackage{hyperref}

\begin{document}

\setcounter{page}{293}
\publyear{2021}
\papernumber{2089}
\volume{183}
\issue{3-4}

  \finalVersionForARXIV

  \title{Automated Repair of Process Models with Non-local Constraints\\ Using State-Based Region Theory}


\author{Anna Kalenkova\thanks{Address for correspondence: School of Computing and Information Systems,
                          The University of Melbourne, Australia}\thanks{This work was partly supported by the Australian Research Council Discovery Project  DP180102839.}~\href{https://orcid.org/0000-0002-5088-7602}{\protect\includegraphics[scale=0.15]{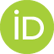}}
\\
School of Computing and Information Systems\\
The University of Melbourne, Australia\\
anna.kalenkova@unimelb.edu.au
\\
\and
            Josep~Carmona\thanks{This work was supported by MINECO and FEDER funds under grant TIN2017-86727-C2-1-R.}~\href{https://orcid.org/0000-0001-9656-254X}{\protect\includegraphics[scale=0.15]{orcid-2.png}}\\
Department of Computer Science\\
Polytechnic University of Catalonia, Spain\\
 jcarmona@cs.upc.edu
\\
\and
    Artem~Polyvyanyy\footnotemark[1]~\href{https://orcid.org/0000-0002-7672-1643}{\protect\includegraphics[scale=0.15]{orcid-2.png}}\\
School of Computing and Information Systems\\
The University of Melbourne, Australia\\
artem.polyvyanyy@unimelb.edu.au
\and
Marcello~La~Rosa\footnotemark[1]~\href{https://orcid.org/0000-0001-9568-4035}{\protect\includegraphics[scale=0.15]{orcid-2.png}}\\
School of Computing and Information Systems\\
The University of Melbourne, Australia\\
marcello.larosa@unimelb.edu.au}

 \maketitle

\runninghead{A.\,Kalenkova et al.}{Automated Repair of  Proc. Models Using Non-local Constr.}

\begin{abstract}
State-of-the-art process discovery methods construct free-choice process models from event logs. Consequently, the constructed models do not take into account indirect dependencies between events. Whenever the input behaviour is not free-choice, these methods fail to provide a precise model. In this paper, we propose a novel approach for enhancing free-choice process models by adding non-free-choice constructs discovered a-posteriori via region-based techniques. This allows us to benefit from the performance of existing process discovery methods and the accuracy of the employed fundamental synthesis techniques.
We prove that the proposed approach preserves fitness with respect to the event log while improving the precision when indirect dependencies exist.
The approach has been implemented and tested on both synthetic and real-life datasets. The results show its effectiveness in repairing models discovered from event logs.
\end{abstract}


\section{Introduction}

\emph{Process mining} is a family of methods used for the analysis of event data, e.g., event logs~\cite{Aalst16}. These methods include \emph{process discovery} aimed at constructing process models from event logs; \emph{conformance checking} applied for finding deviations between real (event logs) and expected (process models) behaviour~\cite{CarmonaDSW18}; and \emph{process enhancement} used for the  enrichment of process models with additional data extracted from  event logs. The latter also includes \emph{process repair} applied to realign process models in accordance with the event logs. Event logs are usually represented as sequences of events (or traces). The main challenge of process discovery is to efficiently construct \emph{fitting} (capturing traces of  the event log), \emph{precise} (not capturing traces not present in the event log) and simple process models.

Scalable process discovery methods, which are most commonly used for the analysis of real-life event data, either produce \emph{directly follows graphs} or use them as an intermediate process representation to obtain a Petri net or a BPMN model~\cite{omg2013bpmn202} (see, e.g., \emph{Inductive miner}~\cite{sander-tree-disc-PN2014} and \emph{Split miner}~\cite{10.1007/s10115-018-1214-x}). Directly follows graphs are 
directed graphs with nodes representing process activities and arcs representing the directly follows (successor) relation between them. Being simple and intuitive, these graphs considerably generalise process behaviour, e.g., they add combinations of process paths that are not observed in the event log. This is because they do not represent higher-level constructs such as parallelism and long-distance (i.e., non-local) dependencies. The above-mentioned discovery methods construct directly follows graphs from event logs and then recursively find relations between sets of nodes in these graphs, in order to discover a {\em free-choice} Petri net~\cite{free-choice}, which can then be seamlessly converted into a BPMN model~\cite{sosym} -- the industry language for representing business process models.
In free-choice nets,
the choice between conflicting activities (such that only one of them can be executed) is always ``free'' from additional  preconditions.
Although free-choice nets can model parallel activities, non-local choice dependencies are modelled by non-free-choice  nets~\cite{10.1007/s10618-007-0065-y}.
Several methods for the discovery of non-free-choice Petri nets exist. However, these methods are either computationally expensive~\cite{10.1007/978-3-540-85758-7_26,van2010process,10.1007/978-3-642-13675-7_14,10.1007/978-3-540-68746-7_24,prime} or heuristic in nature
(i.e., the derived models may fail to replay the traces in the event log)~\cite{10.1007/s10618-007-0065-y}. {\color{black} {Other methods are exact and demonstrate good performance, but they usually produce unstructured process models.~\cite{DBLP:journals/computing/ZelstDAV18,est,DBLP:conf/apn/MannelA19}. In contrast,  the approach proposed in this paper starts with a simple free-choice ``skeleton'', which is then enhanced with additional constraints.}}


In this paper, we propose a repair approach to enhance free-choice nets
by adding extra constructs to capture non-local dependencies. To find non-local dependencies, a transition system constructed from the event log is analysed. This analysis checks whether all the free-choice constructs of the original process model correspond to free-choice relations in the transition system.  For process activities with non-free-choice relation in the transition system but with free-choice relation in the process model, region theory~\cite{BadouelBD15}
is applied to identify, whenever possible, additional places and arcs to be added to the Petri net to ensure the non-local relations between the corresponding transitions. Remarkably, although we have implemented our approach over {\em state-based} region theory~\cite{10.1007/978-3-540-85758-7_26,van2010process,10.1007/978-3-642-13675-7_14}, the proposed approach can also be extended to {\em language-based} region theory~\cite{10.1007/978-3-540-68746-7_24,BergenthumDLM08}, or to geometric or graph-based approaches that have been recently proposed~\cite{BestDS17,SchlachterW18}.

Importantly,
we apply a goal-oriented state-based region algorithm to those parts of the transition system where the free-choice property is not fulfilled. This allows us to reduce the computation time, relegating region-theory to when it is needed. We prove that important quality metrics of the initial free-choice (workflow) net are either preserved or improved for those cases where non-local dependencies exist, i.e., {\em fitness} is never reduced, and {\em precision} can increase. Hence, when using our approach on top of an automated discovery method that returns a free-choice Petri net, one can still keep the complexity of process discovery manageable, obtaining more precise process models that represent the process behaviour recorded in the event log.

In contrast to the existing process repair techniques, which change the structure of the process models by inserting, removing~\cite{10.1145/2980764,10.1007/978-3-319-69462-7_5,FAHLAND2015220} or replacing tasks and sub-processes~\cite{d429471f542a45f09a62131cf704e730}, the approach proposed in this paper only imposes additional restrictions on the process model behaviour, preserving fitness and improving precision where possible.

We implemented the proposed approach as a plugin of Apromore~\cite{la2011apromore}\footnote{https://apromore.org} and tested it both on synthetic and real-world event data. The tests show the effectiveness of our approach within reasonable time bounds.

\medskip
{\color{black}
This article is an extended version of our conference paper~\cite{petrinets2020}. It makes the following additions to the original conference paper:

\begin{compactitem}
\item Extends the repair approach to the set of free-choice models with \emph{silent} transitions;
\item Presents an optimized version of the repair algorithm and analyses its time complexity;
\item Introduces a technique to convert workflow nets with non-local constraints to high-level BPMN models with data objects;
\item Reports on large-scale experiments of applying the proposed techniques \newtext{to real-world data}.
\end{compactitem}
}

\medskip
The paper is organized as follows. Section~\ref{sec:motiv} illustrates the approach by a motivating example.  Section~\ref{sec:prelim} contains the main definitions used throughout the paper. The state-based region technique is introduced in Section~\ref{sec:state}. The proposed model repair approach is then described in Section~\ref{sec:enhanc}. Additionally, Section~\ref{sec:enhanc} contains formal proofs of the properties of the repaired process model. High-level process modelling constructs, e.g., BPMN modelling elements representing non-free-choice routing, are also discussed in Section~\ref{sec:enhanc}. The results of the experiments are presented in Section~\ref{sec:eval}.  Finally, Section~\ref{sec:concl} concludes the paper.


\section{Motivating example}
\label{sec:motiv}
\begin{figure}[h!]
	\vspace{-30pt}
	\centering
	\includegraphics[scale=0.7]{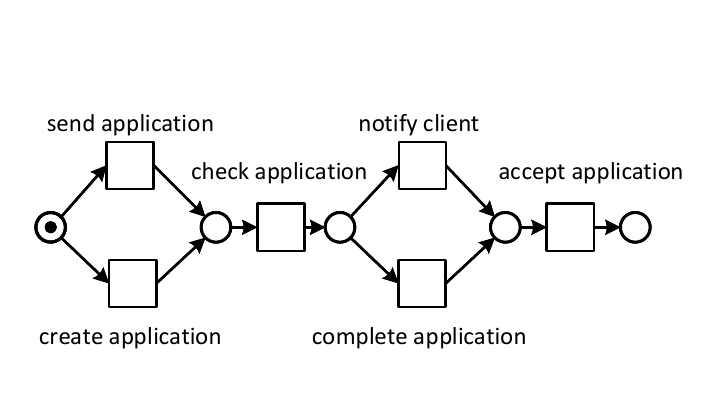}	\vspace{-5mm}
	\caption{A workflow net discovered from $L$ by Inductive miner and Split miner.}
	\label{fig:example}
\end{figure}

This section presents a simple motivating example inspired by the real-life \scalebox{0.9}{BPIC'2017} event log\footnote{https://data.4tu.nl/repository/uuid:5f3067df-f10b-45da-b98b-86ae4c7a310b} and examples discussed in~\cite{10.1007/s10618-007-0065-y}. Consider the process of a loan application.
The process can be carried out by a client or by a bank employee on behalf of the client. Thus, this process can be described by two possible sequences of events (traces) which together can be considered as an event log $L=$$\{\langle\mathit{send\,application},\,\mathit{check\,application},\,\mathit{notify\,client},\,\mathit{accept\,application}\rangle$,
\allowbreak$\langle\mathit{create\,application},$\allowbreak$\mathit{check\,application},\,\mathit{complete\,application},\,\mathit{accept\,application}\rangle\}$.
According to one trace, the client sends a loan application to the bank, then this application is checked. After that, the client is notified, and the application is accepted. The other trace corresponds to a scenario when the application is initially created by a bank employee and checked. After that, the bank employee contacts the client to complete the application. Finally, the application is accepted. Figure~\ref{fig:example} presents a workflow net discovered by Inductive miner~\cite{sander-tree-disc-PN2014} and Split miner~\cite{10.1007/s10115-018-1214-x} from $L$.
This model accepts two additional traces: $\langle\mathit{send\,application},\,\mathit{check\,application},\,\,\mathit{complete\,application},\,\,\mathit{accept\,application}\rangle$,\,\, $\langle\mathit{create\,application},$\allowbreak$\mathit{check\,application},\,\mathit{notify\,client},\,\mathit{accept\,application}\rangle$ not presented in $L$.
These traces violate the business logic of the process. If the application was sent by a client, it is completed, and there is no need to take \newtext{the} $\mathit{complete\,application}$ step. Also, if the application was initially created by a bank employee, the step $\mathit{complete\,application}$ is mandatory.

\begin{figure}[h!]
	\vspace{-10pt}
	\centering
	\includegraphics[scale=1.0]{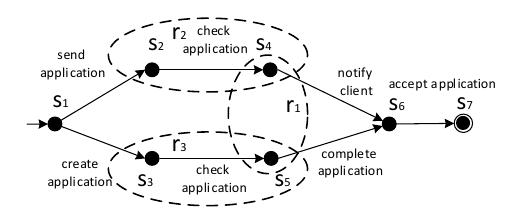}
	\caption{Transition system that encodes event log $L$.}
	\label{fig:example_ts}
\end{figure}

This example demonstrates that the choice between \newtext{the} $\mathit{notify\,client}$ and $\mathit{complete\,application}$ activities depends on the history of the trace. The transition system in Figure~\ref{fig:example_ts} shows the behaviour recorded in event log $L$. State $s_1$ corresponds to a choice between activities $\mathit{send\,application}$ and $\mathit{create\,application}$. This choice does not depend on any additional conditions. In contrast, for the system being in states $s_4$ and $s_5$, there is no free choice between $\mathit{notify\,client}$ and $\mathit{complete}$\allowbreak$\mathit{application}$ activities. In state $s_4$, only \newtext{the} $\mathit{notify\,client}$ activity can be executed, while in state $s_5$, only $\mathit{complete\,application}$ can be performed. This means that there are states in the transition system where \newtext{the} activities  $\mathit{notify\,client}$ and $\mathit{complete\,application}$ are not in a free-choice relation (the choice depends on additional conditions and is predefined), while they are in a free-choice relation in the discovered model (Figure~\ref{fig:example}).

\medskip
To impose additional restrictions on the process model the state-based region theory can be applied~\cite{707587,CarmonaCK10,10.1007/978-3-642-13675-7_14}. Figure~\ref{fig:example_ts} presents three regions $r_1=\{s_4,s_5\}$, $r_2=\{s_2,s_4\}$, and $r_3=\{s_3,s_5\}$ with outgoing transitions labelled by $\mathit{notify\,client}$ and $\mathit{complete\,application}$ events discovered by the state-based region algorithm~\cite{10.1007/978-3-642-13675-7_14}.

\begin{figure}[h!]
	\centering
	\includegraphics[scale=0.7]{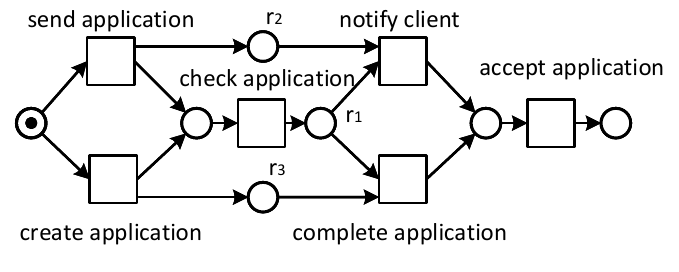}\vspace*{-2mm}
	\caption{A workflow net enhanced with additional regions (places) $r_2$ and $r_3$.}
	\label{fig:example_enhanced}
\end{figure}

Figure~\ref{fig:example_enhanced} presents a target workflow net obtained from the initial workflow net (Figure~\ref{fig:example}) by inserting places that correspond to the discovered regions. As one may note, in addition to $r_1$, two places $r_2$ and $r_3$ were added. These places impose additional constraints, such that the enhanced process model accepts event log $L$ and does not support additional traces and, hence, is more precise.

In the next sections, \newtext{we} describe an approach that implements this idea.

\section{Preliminaries}
\label{sec:prelim}
In this section, we formally define event logs and process models, such as transition systems, Petri nets, and workflow nets.

\subsection{Sets, multisets, event logs}

Let $S$ be a finite set. A \emph{multiset} $m$ over $S$ is a
mapping $m: S\rightarrow\mathbb{N}_0$, where $\mathbb{N}_0$ is the set of all natural
numbers (including zero), i.e., multiset $m$ contains $m(s)$
copies of element $s\in S$.

For two multisets $m,m'$ we write $m\subseteq m'$  iff
$\forall s\in S: m(s) \leq m'(s)$ (the inclusion relation). The
sum of two multisets $m$ and $m'$ is defined as:
$\forall s\in S: (m+m')(s)=m(s)+m'(s)$. The difference of two multisets is a partial function: $\forall s\in S$, such that $m(s)\geq{m(s')}$, $(m-m')(s)=m(s)-m'(s)$.

Let $E$ be a finite set of events. A \emph{trace} $\sigma$ (over $E$)  is a finite sequence of events, i.e., $\sigma\in{E^*}$, where $E^*$ is the set of all finite sequences over $E$, including the empty sequence of zero length. An \emph{event log} $L$ is a set of traces, i.e., $L\subseteq E^*$.

\subsection{Transition systems, Petri nets, workflow nets}

Let $S$  and $E$ be two disjoint non-empty sets of \emph{states} and \emph{events},{\color{black}{ $B\subseteq{S}\times{E_\tau}\times{S}$, where $E_\tau = E\cup \{\tau\}$ and $\tau\notin E$ is a special \emph{silent} event, be a \emph{transition relation}.}}
A \emph{transition system} is a tuple $\TS=(S,E,B,s_{\init}, S_{\fin})$, where  $s_{\init}\in S$ is an initial state and $S_{\fin} \subseteq S$ -- a set of final states.
Elements of $B$ are called  \emph{transitions}. We write $s\tu{e}s'$, when $(s,e,s')\in B$ and
$s\tu{e}$, when $\exists s'\in S$, such that $(s,e,s')\in B$; $s\not\tu{e}$, otherwise. {\color{black}Transition system $\TS$ is \emph{$\tau$-free} iff $\forall (s,e,s')\in B$ it holds that $ e\neq\tau$.}

\medskip
{\color{black}A trace $\sigma= \left\langle e_1,\dots,e_m\right\rangle$ is called \emph{feasible} in $\TS$ iff $\exists  s_1,\dots,s_{n}\in S: \ s_{\init}\tu{\bar e_1}s_1 \tu{\bar e_2}\dots\tu{\bar e_n}s_{n},$ $n\geq m$, $s_n\in S_{\fin}$, and $\left\langle e_1,\dots,e_m\right\rangle=\left\langle \bar e_1,\dots,\bar e_n\right\rangle|_{\{\tau\}}$, where $\left\langle \bar e_1,\dots,\bar e_n\right\rangle|_{\{\tau\}}$ is a sequence obtained from $\left\langle \bar e_1,\dots,\bar e_n\right\rangle$ by removing all $\tau$ without changing the order of the remaining elements, i.e., a \emph{feasible} trace leads from the initial state to some final state possibly taking silent transitions. }A~\emph{language accepted by} $\TS$ is defined as the set of all traces feasible  in $\TS$, and is denoted  \linebreak by $\mathcal{L}(TS)$.

We say that a transition system $\TS$ \emph{encodes} an event log $L$ iff {\color{black}it is $\tau$-free and} each trace from $L$ is a feasible trace in $\TS$, and inversely each feasible trace in $\TS$ belongs to $L$. An example of a {\color{black} $\tau$-free} transition system is shown in Figure~\ref{fig:example_ts}. States and transitions are presented by vertices and directed arcs respectively. The initial state $s_1$ is marked by an additional incoming arrow, while the only final state $s_7$ is indicated by a circle with double border.

\medskip
Let $P$ and $T$ be two finite disjoint sets of \emph{places} and \emph{transitions}, and  $F\subseteq(P\times T)\cup(T\times P)$ be a flow relation.
Let also $E$ be a finite set of events, {\color{black} and $l: {T}\rightarrow{E_\tau}$, where $E_\tau=E\cup \{\tau\}$, $\tau\notin E$, be a labelling function, such that $\forall t_1,t_2\in T, t_1\neq t_2$, $l(t_1)\neq\tau$, it holds that $l(t_1)\neq l(t_2)$, i.e., all the \emph{non-silent} transitions are uniquely labelled.}
Then $N=(P,T,F,l)$ is a \emph{Petri net}.{ \color{black} If $\forall t\in T: l(t)\neq\tau$, then $N$ is a \emph{$\tau$-free} Petri net.}

A \emph{marking} in a Petri net is a multiset over the set of its places. A marked Petri net $(N,m_0)$ is a Petri net $N$ together with its \emph{initial marking} $m_0$.

Graphically, places
are represented by circles, transitions by boxes, and the flow
relation $F$ by directed arcs. Places may carry tokens represented
by filled circles. A current marking $m$ is designated by putting
$m(p)$ tokens into each place $p\in P$. Marked Petri nets are presented in Figures~\ref{fig:example} and~\ref{fig:example_enhanced}.

\medskip
For a transition $t\in T$, an arc $(p,t)$ is called an \emph{input	arc}, and an arc $(t,p)$
an \emph{output arc}, $p\in P$. The \emph{preset} $\pre{t}$ and the
\emph{postset} $\post{t}$ of transition $t$ are defined as the multisets over $P$,
such that $\pre{t}(p)=1$, if $(p,t)\in F$, otherwise $\pre{t}(p)=0$, and $\post{t}(p)= 1$ if $(t,p)\in F$, otherwise $\post{t}(p)=0$.
A transition $t\in T$ is \emph{enabled} in a marking $m$
iff
$\pre{t}\subseteq m$.
An enabled transition $t$
may \emph{fire} yielding  a new marking $m'=_{\mbox{\small def}}
m-\pre{t}+\post{t}$ (denoted $m\tu{t}m'$, $m\tu{l(t)}m'$, or just $m\to m'$).
We say that $m_n$ is \emph{reachable} from $m_1$ iff there
is a (possibly empty) sequence of firings $m_1\to  \dots\to m_n$
and denote
this relation
by $m_1\stackrel{*}{\rightarrow}{m_n}$.

${\mathcal R}(N,m)$ denotes the set of all markings reachable in Petri net $N$ from marking~$m$.
A marked Petri net $(N,m_0),N=(P,T,F,l)$ is \emph{safe} iff $\forall p\in P,\forall m\in\mathcal{R}(N,m_0):m(p)\leq 1$, i.e., at most one token can appear in a place.

\medskip
A \emph{reachability graph} of a marked Petri net $(N,m_0)$, $N=(P,T,F,l)$, with a labelling function {\color{black}$l:T\rightarrow E_\tau$}, is a
transition system $\TS=(S,E,B,s_{\init}, S_{\fin})$ with the set of states $S= {\mathcal R}(N,m_0)$ and transition relation $B$ defined by $(m,e,m')\in B$ iff $m\tu{t}m'$, where $e=l(t)$. The initial state in $\TS$ is the initial marking $m_0$. If some reachable markings in $(N,m_0)$ are distinguished as final markings, they are defined as final states in $\TS$. The \emph{language} of a Petri net $(N,m_0)$, denoted by $\mathcal{L}(N,m_0)$ is the language of its reachability graph, i.e., $\mathcal{L}(N,m_0)=\mathcal{L}(\TS)$. We say that a Petri net $(N,m_0)$ \emph{accepts} a trace iff  this trace is feasible in the reachability graph of  $(N,m_0)$; a Petri net \emph{accepts} a language iff  this language  is accepted by its reachability graph. {\color{black} When $S$ is finite we can construct a $\tau$-free transition system $\widehat{\TS}$, such that $\mathcal{L}(\widehat\TS)=\mathcal{L}(\TS)$~\cite{hopcroft}, we will call it a \emph{$\tau$-closure} of the reachability graph $\TS$.}

\begin{figure}[h]
	\vspace{-2mm}
	\centering
	\includegraphics[scale=0.9]{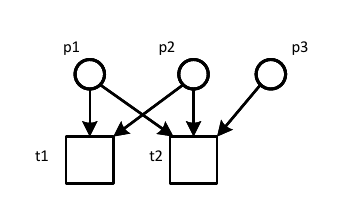}	\vspace{-5pt}
	\caption{A non-free-choice Petri net.}
	\label{fig:non-free-choice}
\end{figure}

\medskip
Given a Petri net $N=(P,T,F,l)$, two transitions  $t_1,t_2\in T$ are in a \emph{free-choice relation} iff $\pre{t_1}\cap\pre{t_2}=\emptyset$ or $\pre{t_1}=\pre{t_2}$. {\color{black}{When $l(t_1)\neq\tau$ and $l(t_2)\neq\tau$,}} we also say that events (or activities) $l(t_1)$ and $l(t_2)$ are in a \emph{free-choice relation}. Petri net $N$ is called \emph{free-choice} iff for all $t_1,t_2\in T$, it holds that $t_1$ and $t_2$ are in a free-choice relation. This is one of the several equivalent definitions for free-choice Petri nets presented in~\cite{free-choice}. A Petri net is called \emph{non-free-choice} iff it is not free-choice. Figure~\ref{fig:non-free-choice} presents an example of a non-free-choice Petri net, where for two transitions $t_1$ and $t_2$ \newtext{it} holds that $\pre{t_1}\cap\pre{t_2}=\{p_1,p_2\}\neq\emptyset$ and $\pre{t_1}=\{p_1,p_2\}\neq\pre{t_2}=\{p_1,p_2,p_3\}$.

\medskip
The choice of which transition will fire depends on an additional constraint imposed by place $p_3$. If $m(p_1)>0$, $m(p_2)>0$, and $m(p_3)=0$, then only $t_1$ is enabled, thus there is no free-choice between $t_1$ and $t_2$. Another example of a non-free-choice Petri net was presented earlier in Figure~\ref{fig:example_enhanced}, where transitions labelled by $\mathit{notify\,client}$ and $\mathit{complete\,application}$ are not in a free-choice relation, thus the Petri net is not free-choice. An example of a free-choice Petri net is presented in Figure~\ref{fig:example}.

\medskip
Workflow nets \newtext{are} a special subclass of Petri nets designed for modelling workflow processes~\cite{10.1007/3-540-47961-9_37}.
A workflow net has one initial and one final place, and every place or transition is on a directed path from the initial to the final place.

\medskip
Formally, a marked Petri net $N=(P,T,F,l)$ is called a  \emph{workflow net} iff

\begin{enumerate}
	\item
	There is one source place  $i\in P$ and one sink place $o\in P$, such that $i$ has no input arcs and $o$ has no output arcs.
	\item
	Every node from $P\cup T$ is on a directed path from  $i$ to $o$.
	{\color{black}
	\item
	The initial marking contains one token in the source place.
	\item
	The final markings contain one token in the sink place.
	}
\end{enumerate}
The \emph{language} of \newtext{a} workflow net $N$ is denoted by $\mathcal{L}(N)$.

\medskip
{\color{black} A workflow net $N$ with the initial marking $[i]$ containing only one token in the source place and the final marking $[o]$ containing only one token in the sink place is \emph{sound} iff}

\begin{enumerate}
	\item
	For every state $m$ reachable in $N$, there exists a firing sequence leading from $m$ to the final state~$[o]$. Formally, $\forall m:[([i]\stackrel{*}{\rightarrow}m)$ implies $(m\stackrel{*}{\rightarrow}[o])]$;
	\item
	The state $[o]$ is the only state reachable from $[i]$ in $N$  with at least one token in place~$o$. Formally, $\forall m:[([i]\stackrel{*}{\rightarrow}m)\wedge ([o]\subseteq m)$ implies $(m=[o])]$;
	\item
	There are no \emph{dead} transitions in $N$. Formally, $\forall t\in T \ \exists m,m':([i]\stackrel{*}{\rightarrow}m\stackrel{t}{\rightarrow}m')$.
\end{enumerate}
Note that both models presented in Figures~\ref{fig:example} and~\ref{fig:example_enhanced} are sound workflow nets.

\section{Region state-based synthesis}
\label{sec:state}

In this section, we give a brief description of the well-known state-based region algorithm~\cite{707587} applied for the synthesis of Petri nets from transition systems.

\medskip
Let $\TS=(S,E,T,s_{\init}, S_{\fin})$ {\color{black}be a $\tau$-free transition system with a finite set of states $S$} and ${r}\subseteq{S}$ be a subset of states. Subset $r$ is a \emph{region} iff for each event ${e}\in{E}$ one of the following conditions holds:
\begin{itemize}
	\item all the transitions ${s_1}\stackrel{e}{\rightarrow}{s_2}$ \emph{enter} $r$, i.e., ${s_1}\notin{r}$ and ${s_2}\in{r}$,
	\item all the transitions ${s_1}\stackrel{e}{\rightarrow}{s_2}$ \emph{exit} $r$, i.e., ${s_1}\in{r}$ and ${s_2}\notin{r}$,
	\item all the transitions ${s_1}\stackrel{e}{\rightarrow}{s_2}$ \emph{do not cross} $r$, i.e., ${s_1},{s_2}\in{r}$ or ${s_1},{s_2}\notin{r}$.
\end{itemize}

In other words, all the transitions labelled by the same event are of the same type (\emph{enter}, \emph{exit}, or \emph{do not cross}) for a particular region.

A region $r'$ is said to be a \emph{subregion} of a region $r$ iff ${r'}\subseteq{r}$. A region $r$ is called a \emph{minimal region} iff it does not have \newtext{any subregions other than $r$}.

\begin{figure}[h!]
	\begin{center}
		\includegraphics[width=0.65\textwidth]{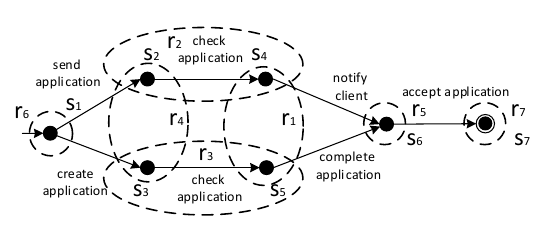}
	\end{center}\vspace*{-5mm}
	\caption{Applying the state-based region algorithm to the transition system presented in Figure~\ref{fig:example_ts}.}
	\label{fig:all-regions}
\end{figure}

\begin{figure}[h!]
	\begin{center}
		\includegraphics[width=0.65\textwidth]{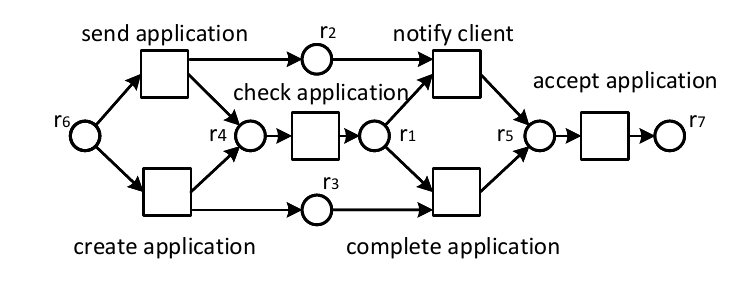}
	\end{center}\vspace*{-7mm}
	\caption{A Petri net synthesized from the transition system presented in~Figure~\ref{fig:all-regions}.}
	\label{fig:final_model}\vspace{-6mm}
\end{figure}

\medskip
The state-based region algorithm covers the transition system  by its minimal regions~\cite{DeselR96}. Figure~\ref{fig:all-regions} presents the transition system from Figure~\ref{fig:example_ts} covered by minimal regions: $r_1=\{s_4,s_5\}$, $r_2=\{s_2,s_4\}$, $r_3=\{s_3,s_5\}$, $r_4=\{s_2,s_3\}$, $r_5=\{s_6\}$, $r_6=\{s_1\}$, and $r_7=\{s_7\}$.
According to the algorithm in~\cite{707587}, every minimal region is transformed to a place $p$ in the target Petri net and connected with transitions corresponding to the \emph{exiting} and \emph{entering} events by outgoing and incoming arcs, respectively (refer to Figure~\ref{fig:final_model}). {\color{black} If a region contains the initial state of $\TS$, the corresponding place $p$ is added to the initial marking of the target Petri net. If a region contains a final state of $\TS$, new final markings are obtained by adding $p$ to the existing final markings and these new final markings are added to the overall set of the final markings of the target Petri net.}

Region $r$ {\em separates} two different states $s,s'\in S$, $s\neq s'$, iff  $s \in r$ and $s' \notin r$.  Finding such a region is the {\em state separation problem} between $s$ and $s'$ and is denoted by $\SSP(s,s')$.  When an event $e$ is not enabled in a state $s$, i.e., ${s}\not\tu{e}$, a region $r$, containing $s$ may be found, such that $e$ does not \emph{exit} $r$. Finding such a region is known as the {\em event/state separation problem} between $s$ and $e$ and is denoted by $\ESSP(s,e)$.

A well-known result in region theory establishes that if all $\SSP$ and $\ESSP$ problems are solved, then synthesis is exact~\cite{BadouelBD15}:
\begin{theorem} A {\color{black} $\tau$-free $TS$ with a finite number of states} can be synthesized into a safe {\color{black} $\tau$-free} Petri net $N$ with the reachability graph isomorphic to $TS$ if all $\SSP$ and $\ESSP$ problems are solvable.
\end{theorem}

These problems are also known to be NP-complete~\cite{BadouelBD15}. In this paper, we reduce the size of these problems by constructing regions corresponding to particular events only.

\section{Repairing free-choice process models}
\label{sec:enhanc}
In this section, we describe our approach for repairing free-choice workflow nets using non-local constraints captured in the event logs. Additionally, we investigate  formal properties of the repaired process models.

\subsection{Problem definition}
Let $N$ be a free-choice workflow net discovered from event log $L$ and
let $\TS$ be a transition system encoding $L$.
Due to limitations of the automated discovery methods~\cite{sander-tree-disc-PN2014,10.1007/s10115-018-1214-x}   that construct free-choice workflow nets {\color{black}(even if they discover models from a wider class of workflow nets with silent transitions)}, not all the places that correspond to minimal regions may have been derived, and therefore important $\SSP$/$\ESSP$ problems may not be solved in $N$, when considering {\color{black}$\tau$-closure of the reachability graph $\mathcal{\widehat{R}}(N,[i])$} as the behaviour to represent with $N$.

\medskip
This brings us to the following characterization of the problem.
Let $t_1, \ldots, t_n$ be {\color{black} non-silent transitions, i.e., $l(t_1)\neq\tau, \ldots, l(t_n)\neq\tau$,} in $N$ with $\pre{t_1}=\pre{t_2}=\dots=\pre{t_n}$, i.e.,  $t_1, \ldots, t_n$ are in the free-choice relation in $N$, and let $\TS=(S,E,T,s_{\init}, S_{\fin})$ be a minimal transition system encoding the event log $L$. If there exists a state $s \in S$, and $1 \le i < j \le n$ such that:
\begin{enumerate}
\itemsep=0.95pt
	\item $e_i$, $e_j$ correspond to transitions $t_i$, $t_j$, respectively,
	\item ${s}\stackrel{e_i}{\rightarrow}$,
	\item ${s}\not\stackrel{e_j}{\rightarrow}$
\end{enumerate}
\noindent Then, the relation of $t_1, \ldots, t_n$ in $N$ corresponds to a \emph{false free-choice relation}, not observed in $\TS$.

\eject
There is no place in $N$ corresponding to a region that solves the $\ESSP(s,e_j)$ problem, because  $t_1, \ldots, t_n$ are in a free-choice relation in $N$.
For instance, the Petri net in Figure~\ref{fig:example} contains places corresponding to regions $r_1$, $r_4$, $r_5$, $r_6$, and $r_7$ shown in Figure~\ref{fig:all-regions}, and none of those regions solves the  $\ESSP(s_4,\mathit{complete\,application})$ and $\ESSP(s_5,\mathit{notify\,client})$ problems in the transition system.

Note that we define the notion of a false free-choice relation for a minimal
transition system (transition system with a minimal number of states~\cite{hopcroft}) encoding the event log. This is done in order to avoid the case when there exists a state $s'$ which is equivalent to $s$,
such that ${s'}\stackrel{e_j}{\rightarrow}$. During the minimization, these equivalent states will be merged into one state with outgoing transitions labelled by $e_i$ and $e_j$ showing that there is no false free-choice relation between corresponding transitions. Another reason to minimize the transition system is to reduce the number of states being analysed.

Note that there is no guarantee that an $\ESSP$ problem can be solved. Nevertheless, in the running example, regions $r_2$ and $r_3$ solve $\ESSP(s_4,\mathit{complete\,application})$ and $\ESSP(s_5,\mathit{notify\,client})$ problems.

\subsection{Algorithm description}

In this subsection, we present an algorithm for  enhancement of a free-choice workflow net $N$ with additional constraints from event log $L$ (Algorithm~\ref{alg:distance}).
\begin{algorithm}[h!]
	\DontPrintSemicolon
	\SetAlgoLined
	\SetNoFillComment
	\caption{RepairFreeChoiceWorkflowNet \label{alg:distance}}
	\KwIn{Free-choice workflow net $N$; Event log $L$.}
	\KwOut{Repaired net $N'$ obtained from $N$ by inserting additional non-local constraints.}
	\tcc*[l]{Construct minimal transition system}
	$\mathit{TS}\leftarrow\textrm{ConstructMinTS(L);}$
	\BlankLine
	\tcc*[l]{Compute ESSP problems}
	$\mathit{ESSPProblems}\leftarrow\textrm{FindFalseFreeChoiceRelations(N,TS);}$
	\BlankLine
	$\mathit{N'}\leftarrow\mathit{N};$
	\BlankLine
	\ForEach{$(s,e)\, from\, \mathit{ESSPProblems}$}
	{
		\tcc*[l]{Solve $\mathit{ESSP}(s,e)$}
		$\mathit{Y}\leftarrow\textrm{ComputeRegionsESSP(TS,s,e);}$
		\BlankLine
		\uIf{($\mathit{Y} \ne \emptyset$)}{
			\tcc*[l]{$\mathit{ESSP}(s,e)$ has been solved }
			$\mathit{N'}\leftarrow\textrm{AddNewConstraints}(\mathit{N',Y});$}
	}
	\Return{$N'$}\;
\end{algorithm}

Firstly, by applying $\mathit{ConstructMinTS}$, a minimal transitional system encoding the event log $L$ is constructed.\footnote{Transition system can be constructed from the event log as a prefix-tree~\cite{van2010process} with subsequent minimization~\cite{hopcroft}.} Then, false free-choice relations and corresponding $\ESSP$ problems are identified. According to the definition of a false free-choice relation presented earlier, \newtext{the} procedure for finding false free-choice relations $\mathit{FindFalseFreeChoiceRelations}$  is polynomial in time. Indeed, to find all the false free-choice relations one needs to check  whether all the states of the transition system have none or all outgoing transitions labelled by events assumed to be in free-choice relations within the original workflow net $N$. When the false free-choice relations are discovered, for each corresponding $\ESSP$ problem, \newtext{the} function $\mathit{ComputeRegionsESSP}$, which finds regions solving the $\ESSP$ problem, is applied. {\color{black} Finally, if new regions solving  $\ESSP$ problems are found, the function $\emph{AddNewConstraints}$ adds the corresponding constraints (places) to the target workflow net $N'$.

Since the problem of finding minimal regions which solve \newtext{the} $\ESSP$ problem is known to be NP-complete, this is the most time consuming step of Algorithm~\ref{alg:distance}.

\medskip
Suppose that transitions $t_1, \ldots, t_n$  in $N$ labelled by $e_1, \ldots, e_n$, respectively, are in a false free-choice relation. Hence, there exist a state $s$ in $\TS$ and two events $e_i$, $e_j$, $i,j\in[1,\ldots,n]$, such that ${s}\stackrel{e_i}{\rightarrow}$ and ${s}\not\stackrel{e_j}{\rightarrow}$. To solve $\ESSP(s,e_j)$ problem we need to find a region $r$, such that $s\in r$ and $e_j$ does not \emph{exit} $r$ (Figure~\ref{fig:essp1}). Then, this region will be converted to a place (Figure~\ref{fig:essp2}) that imposes additional behavioural constraints. Note that region $r'$ in Figure~\ref{fig:essp1} and Figure~\ref{fig:essp2} corresponds to the original free-choice relation.

\begin{figure*}[h!]
	\label{fig:ex}
	\centering
	\begin{subfigure}{.45\textwidth}
		\begin{center}
			\includegraphics[scale=1.2]{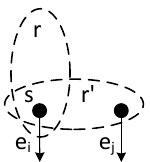}
		\end{center}
		\caption{Solving $\ESSP(s,e_j)$ problem.}
		\label{fig:essp1}
	\end{subfigure}
	\quad
	\begin{subfigure}{.45\textwidth}
		\vspace{14pt}
		\begin{center}
			\includegraphics[scale=1.1]{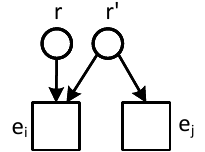}
		\end{center}
		\vspace{-2pt}
		\caption{Adding non-local constraint $r$.}
		\label{fig:essp2}
	\end{subfigure}\vspace*{-1mm}
	\caption{\newtext{Finding a new region that solves the $\ESSP$ problem.}}
\end{figure*}

The goal of the repair algorithm is to find additional constraints for the transitions that are in false free-choice relations. Therefore, we may narrow the search space and consider only those regions containing $s$ where $e_i$ is the exiting event.  Since $e_i$ may be involved in several $\ESSP$ problems (with different states $s$ and not-exiting transitions $e_j$), the general approach could be to construct all the minimal regions with the exiting event $e_i$ and check whether these regions do not correspond to the free-choice relation, i.e., not all the events from $\{e_1, \ldots, e_n\}$ are exiting.

According to~\cite{deriving}, when constructing regions, the space of potential solutions expands in no more than two directions for each of the events. Suppose that $|E|$ is a set of events, then the time complexity of constructing regions with exiting event $e_i\in E$ can be estimated as $O(2^{|E|-1})$. Suppose that $E'\subseteq E$ is a set of events corresponding to transitions in false free-choice relations, then the overall time complexity of the repair algorithm is $O(|E'|\cdot 2^{|E|-1})$. Although the upper time bound of the algorithm is exponential, the search space is not always expanded in two directions, and often only one search direction is possible,  or no search directions are possible at all. Moreover, we do not construct all the regions covering the transition system, but only those that repair the model.
In Section~\ref{sec:eval}, we demonstrate that this algorithm can be efficiently applied to repair process models discovered from real-life event data.

}

\subsection{Formal properties}

In this subsection, we prove the formal properties of Algorithm~\ref{alg:distance}. Firstly, we study the relation between the languages of the initial and target workflow nets.
Theorem~\ref{theorem:fitness} proves that if a trace \emph{fits} the initial model (i.e., the initial model accepts the trace), it also \emph{fits} the target model. Although the proof seems trivial, we consider different cases {\color{black}in order to verify that a final marking is reached.}

\begin{theorem}[Fitness]
	\label{theorem:fitness}
	Let $\sigma\in L$ be a trace of an event log $L\in E^*$, and $N=(P,T,F,l)$, {\color{black}$l:T\rightarrow E_{\tau}$} be a free-choice workflow net, such that its language contains $\sigma$, i.e., $\sigma\in\mathcal{L}(N)$.
	Workflow net $N'=(P\cup P',T,F',l)$,  {\color{black}$l:T\rightarrow E_{\tau}$}, is obtained from $N$ and $L$ using Algorithm~\ref{alg:distance}. Then the language of $N'$ contains $\sigma$, i.e., $\sigma\in\mathcal{L}(N')$.
	
	\begin{proof}
		Let us prove that an insertion of a single place by Algorithm~\ref{alg:distance} preserves the ability of the workflow net to accept trace $\sigma$. Consider a place $r$ (Figure~\ref{fig:theorems_b}) constructed from the corresponding region $r$ (Figure~\ref{fig:theorems_a}) with entering events $b_1,\ldots,b_m$ and exiting events $a_1,\ldots,a_q,\ldots a_p$. Events {\color{black}$a_q,\ldots,a_p$} can belong to a larger set of events {\color{black}$a_q,\ldots,a_p,\ldots, a_k$} which are in a free-choice relation within $N$. 
		Consider the workflow net $N'$ with a new place $r$ (the fragment of $N'$ is presented in Figure~\ref{fig:theorems_b}) and \newtext{the} following four cases:
      \begin{figure}[!h]
      \vspace*{-3mm}
			\label{fig:ex}
		\centering
			\begin{subfigure}{.45\textwidth}
				\begin{center}
			\includegraphics[scale=0.8]{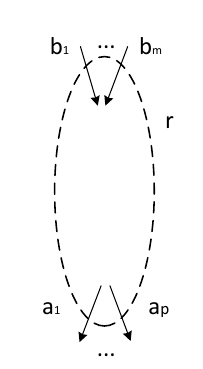}
				\end{center}\vspace{-5mm}
				\caption{A fragment of transition system encoding $L$.}\label{fig:theorems_a}
			\end{subfigure}
			\begin{subfigure}{.45\textwidth}
				\vspace{16pt}
				\begin{center}
 		\includegraphics[scale=0.8]{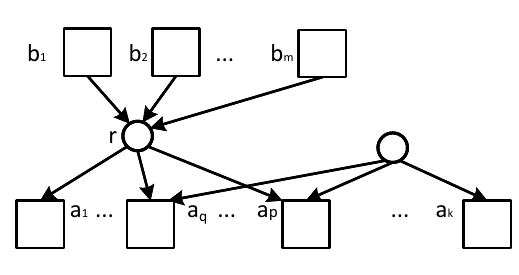}
				\end{center}	\vspace{-2mm}
				\caption{A fragment of $N'$.}\label{fig:theorems_b}
			\end{subfigure}\vspace*{-2mm}
			\caption{Adding a new place $r$.}\vspace*{-4mm}
		\end{figure}

 \begin{enumerate}
			\item Suppose $\sigma=\langle e_1,\ldots,e_l\rangle\in L$ does not contain events from $\{b_1,\ldots,b_m\}$ and $\{a_1,\ldots,$\allowbreak$a_p\}$ sets. Since $\sigma\in\mathcal{L}(N)$, there is a sequence of firings in $N$:{ \color{black} $m_0\stackrel{e_1}{\rightarrow}m_1\stackrel{e_2}{\rightarrow}\ldots\stackrel{e_l}{\rightarrow}m_n$, where $m_0$ and $m_n$ are the initial and final markings of the workflow net respectively.} The same sequence of firings can be repeated within the target workflow net $N'$, because $\sigma$ does not contain events from the sets $\{b_1,\ldots,b_m\}$ and $\{a_1,\ldots,a_p\}$, and the place $r$ is not involved in this sequence of firings.				
			
			\item Now let us consider trace $\sigma=\langle e_1,\ldots,b_i,\ldots, a_j,\ldots,e_l\rangle$ in which each occurrence of event $b_i$ from the set $\{b_1,\ldots,b_m\}$ is followed by an occurrence of event $a_j$ from $\{a_1,\ldots,a_p\}$. Similarly, for the firing sequence within $N$:{ \color{black} $\smash{m_0\stackrel{e_1}{\rightarrow}m_1\stackrel{e_2}{\rightarrow}\ldots  \stackrel{b_i}{\rightarrow}m_i}\rightarrow\smash{\ldots\rightarrow m_j\stackrel{a_j}{\rightarrow}\ldots\stackrel{e_l}{\rightarrow}m_n}$}, there is a corresponding sequence {\color{black}$m_0\stackrel{e_1}{\rightarrow}m_1\stackrel{e_2}{\rightarrow}\ldots  \stackrel{b_i}{\rightarrow}m'_i\rightarrow\ldots\rightarrow m'_j\stackrel{a_j}{\rightarrow}\ldots\stackrel{e_l}{\rightarrow}m_n$} for $N'$, such that $\forall p\in P: m'_i(p)=m_i(p)$, $m'_i(r)=1$, $\forall p\in P: m'_j(p)=m_j(p)$, and $m'_j(r)=1$.\vspace*{-1mm}
			
					\item Consider trace $\sigma$ where an event from  $\{b_1,\ldots,b_m\}$ is not followed by an event from the set $\{a_1,\ldots,a_p\}$. More precisely, there are two possible cases: (1) trace $\sigma$ contains an event from $\{b_1,\ldots,b_m\}$ and does not contain an event from $\{a_1,\ldots,a_p\}$; (2) an occurrence of an event from  set $\{b_1,\ldots,b_m\}$ is followed by another occurrence of an event from the same set $\{b_1,\ldots,b_m\}$ and only after that an event from the set $\{a_1,\ldots,a_p\}$ may follow. For the case (1), it is possible that the final state $s_o$ belongs to the region $r$ (Figure~\ref{fig:theorems2_a}). 	 {\color{black} Then, the firing sequence in N: $\smash{m_0\stackrel{e_1}{\rightarrow}m_1\stackrel{e_2}{\rightarrow}\ldots  \stackrel{b_i}{\rightarrow}m_i}\rightarrow\ldots\stackrel{e_l}{\rightarrow}m_n$ corresponds to the firing sequence {\color{black}$\smash{m_0\stackrel{e_1}{\rightarrow}m_1\stackrel{e_2}{\rightarrow}\ldots  \stackrel{b_i}{\rightarrow}m'_i}\rightarrow\ldots\stackrel{e_l}{\rightarrow}m'_n$} in  $N'$, where $\forall p\in P: m'_i(p)=m_i(p), m'_i(r)=1,$ $\ldots,m'_n(p)=m_n(p),m'_n(r)=1$ and, according to the synthesis algorithm (Section~\ref{sec:state}), $m'_n$ is a new added final marking.}
			
				\begin{figure*}[h!]
				\label{fig:ex}
				\vspace{-3mm}
				\centering
				\begin{subfigure}{.48\textwidth}
					\begin{center}
						\includegraphics[scale=1.0]{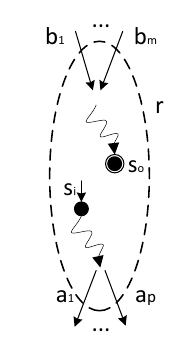}
					\end{center}\vspace{-4mm}
					\caption{Start/final state inside the region $r$.}
					\label{fig:theorems2_a}
				\end{subfigure}
				\begin{subfigure}{.48\textwidth}
					\vspace{20pt}
					\begin{center}
						\vspace{-15pt}
						\includegraphics[scale=1.0]{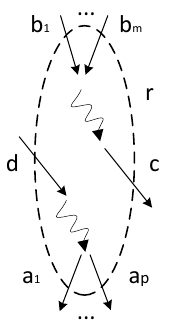}
					\end{center} \vspace{-4mm}
					\caption{The bypass incoming/outgoing transitions.}
					\label{fig:theorems2_b}
				\end{subfigure}
				\caption{Fragments of the transition system that encodes $L$.}
			\end{figure*}

			The other possible scenario for the cases (1) and (2), is that the trace  $\sigma$ does not terminate inside region $r$.
			In both cases, there is a transition labelled by an event $c\notin\{a_1,\ldots,a_p\}$ which exits region $r$ (\newtext{Figure~\ref{fig:theorems2_b}}). While it is obvious for the case (1), for the case (2) this can be proven by the fact that there are two occurrences of events from $\{b_1,\ldots,b_m\}$ with no occurrences of events from $\{a_1,\ldots,a_p\}$ in between, and hence the trace $\sigma$ leaves the region $r$ in order to enter it again with a transition labelled by an event from $\{b_1,\ldots,b_m\}$. Having a new exiting event $c\notin\{a_1,\ldots,a_p\}$ contradicts the definition of the region $r$ which has $\{a_1,\ldots,a_p\}$ as a set of exiting events.
			Thus, we have proven that there is no such a trace in the initial event log containing an event from the set $\{b_1,\ldots,b_m\}$ which is not followed by an event from the set $\{a_1,\ldots,a_p\}$.

			\item Consider the last possible case when an event from $\{a_1,\ldots,a_p\}$ is not preceded by an event from $\{b_1,\ldots,b_m\}$ in trace $\sigma$. Here again we can distinguish two situations: (1) $\sigma$ contains an event from $\{a_1,\ldots,a_p\}$ and does not contain an event from $\{b_1,\ldots,b_m\}$; (2) the occurrence of an event from $\{a_1,\ldots,a_p\}$ is firstly preceded by another occurrence of an event from $\{a_1,\ldots,a_p\}$  which in its turn can be preceded by an event from $\{b_1,\ldots,b_m\}$. Just like in the previous case, two scenarios are possible: the trace starts inside the region~$r$ (Figure~\ref{fig:theorems2_a}.) or there is a transition entering $r$ and labelled by an event $d\notin\{b_1,\ldots,b_m\}$ (Figure~\ref{fig:theorems2_b}). {\color{black}Similarly to the case (3), we can prove that  either (1) we adjust the initial marking, or (2) $r$ is not \linebreak a region.}
\end{enumerate}
		Thus, we have proved that if a place corresponding to a region constructed by the Algorithm~\ref{alg:distance} is added to the initial workflow net $N$ then all the traces \newtext{from} $L$  accepted by $N$ are also accepted by the resulting workflow net $N'$.
\end{proof}
\end{theorem}

The following theorem states that the resulting model cannot be less precise than the initial process model, i.e., it cannot accept new traces which were not accepted by the initial model.

\begin{theorem}[Precision]
	Let $N=(P,T,F,l)$, {\color{black}$l:T\rightarrow E_{\tau}$}, be a free-choice workflow net and let $L$ be an event log over \newtext{a} set of events $E$.
	If workflow net $N'$ is obtained from $N$ and $L$ by Algorithm~\ref{alg:distance}, then the language of $N$  contains the language of $N'$, i.e., $\mathcal{L}(N')\subseteq\mathcal{L}(N)$.
	\begin{proof}
		The proof follows from the well-known result that \newtext{the} addition of new places (preconditions) can only restrict the
		behaviour and, hence, the language of the Petri net~\cite{10.5555/3405}.
	\end{proof}
\end{theorem}

Next, we formulate and prove a sufficient condition for the soundness of the resulting workflow net.
This condition is formulated in terms of the state-based region theory.

\begin{theorem}[Soundness]
	\label{theorem:soundness}
	Let $L$ be an event log over set $E$. Let $N=(P,T,F,l)$, {\color{black}$l:T\rightarrow E_{\tau}$} be a sound free-choice workflow net {\color{black} with initial and final markings $[i]$ and $[o]$, respectively}. Suppose that workflow net $N'=(P\cup P',T,F',l)$ is obtained from $N$ and $L$ by applying {\color{black}Algorithm~\ref{alg:distance} to a set of transitions $t_1,\ldots,t_q$ in a free-choice relation within $N$, such that these transitions are all not silent, i.e., $\forall j\in[1,q]:l(t_j)\neq\tau$.}  Suppose also that $\{r^{(1)},\ldots,r^{(n)}\}$ is a set of regions constructed by Algorithm~\ref{alg:distance} in the transition system encoding $L$ (Figure~\ref{fig:theorems3_b}). Let $E_{\ent}^{(1)}=\{b_1^{(1)},\ldots,b_m^{(1)}\},\ldots,E_{\ent}^{(n)}=\{b_1^{(n)},\ldots,b_t^{(n)}\}$ and $E_{\exit}^{(1)}=\{a_1^{(1)},\ldots,a_p^{(1)}\},\ldots,E_{\exit}^{(n)}=\{a_1^{(n)},\ldots,a_k^{(n)}\}$ be sets of entering and exiting events for the regions $r^{(1)},\ldots,r^{(n)}$ respectively. Consider unions of these sets: $E_{\ent}=E_{\ent}^{(1)}\cup\ldots\cup E_{\ent}^{(n)}$ and $E_{\exit}=E_{\exit}^{(1)}\cup\ldots\cup E_{\exit}^{(n)}$. {\color{black}Suppose that $E_{\exit}=\{l(t_1),\ldots,l(t_q)\}$ and $\forall j,k, j\neq k$ holds that $E_{\exit}^{(j)}\cap E_{\exit}^{(k)}=\emptyset$.} If there exists a (not necessarily minimal) region $r$ in the {\color{black}$\tau$-closure of the} reachability graph of $N$ (Figure~\ref{fig:theorems3_a}) with entering and exiting sets of events $E_{\ent}$ and $E_{\exit}$, respectively, which  does not contain states corresponding to $[i]$ (initial) and $[o]$ (final) markings of $N$, then $N'$ is sound.

	\begin{figure*}[h!]
	\label{fig:theorems3}
	\vspace{-8mm}
	\centering
	\begin{subfigure}{.41\textwidth}
		\begin{center}
			\includegraphics[scale=1.0]{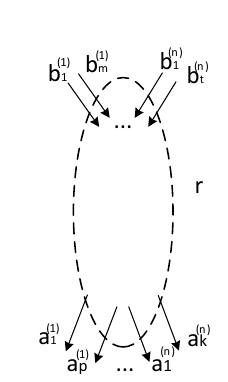}
		\end{center}\vspace{-4mm}
		\caption{A fragment of the reachability graph of $N$.}
			\label{fig:theorems3_a}
	\end{subfigure}
		\begin{subfigure}{.58\textwidth}
		\vspace{56pt}
		\begin{center}
			\vspace{-6pt}
			\includegraphics[scale=0.8]{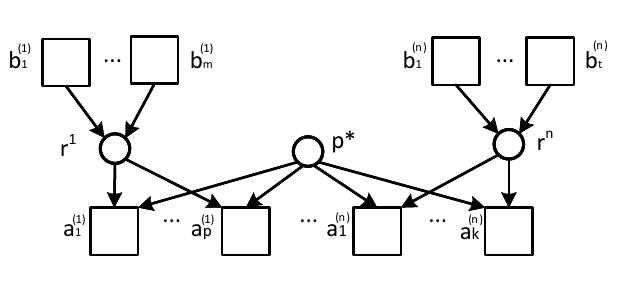}
		\end{center}	\vspace{1mm}
		\caption{A fragment of $N'$.}
				\label{fig:theorems3_b}
	\end{subfigure}
	\caption{Adding new places to $N$.}\vspace*{-4mm}
\end{figure*}
	
	\begin{proof}
		Repeating the proof of Theorem~\ref{theorem:fitness} and taking into account that the initial and final states of {\color{black}$\tau$-closure of} the reachability graph of $N$ do not belong to the region $r$, we can state that there is a following relation between  $E_{\ent}$ and $E_{\exit}$ within $\mathcal{L}(N)$, i.e, for each trace, each occurrence of the event from $E_{\ent}$ is followed by an occurrence of the event from $E_{\exit}$ and there are no other occurrences of events from $E_{\ent}$ between them.
		
		The firing sequences of $N'$ which do not involve firings of transitions labelled by events from $E_{\ent}$ and $E_{\exit}$ repeat the corresponding firing sequences of $N$ and do not violate the soundness of the model.

		Let us consider a firing sequence of $N'$ which involves firings of transitions labelled by events  from  $E_{\ent}$ and $E_{\exit}$. Consider $b\in E_{\ent}$, the firing sequence enabling and firing $b$ in $N'$: $[i]\stackrel{*}{\rightarrow}m'_1\stackrel{b}{\rightarrow}m'_2$,  corresponds to the firing sequence performed by $N$: $[i]\stackrel{*}{\rightarrow}m_1\stackrel{b}{\rightarrow}m_2$, where $\forall p\in P: m_1(p)=m'_1(p)$, $m_2(p)=m'_2(p)$. Without loss of generality, suppose that $b\in E^{(i)}_{\ent}$, then $m'_2(r^i)=1$, where  $r^i$ is a place constructed by Algorithm~\ref{alg:distance}.
		
		Since $E_{\ent}$ and $E_{\exit}$ events are in a following relation within $\mathcal{L}(N)$, they are in the following relation within $\mathcal{L}(N')$, because $\mathcal{L}(N')\subseteq \mathcal{L}(N)$. Consider sequences of steps leading to some of the events from $E_{\exit}$. These firing sequences will be: $m'_2\stackrel{*}{\rightarrow}m'_3$ and
		$m_2\stackrel{*}{\rightarrow}m_3$, where $m_3(p)=m'_3(p)$ and $m'_3(r^i)=1$, in $N'$ and $N$ respectively.
		
		In model $N'$  transitions labelled by the events from $E^{(i)}_{\exit}$ will be enabled in $m'_3$, all other trans\-itions labelled by events from $E_{\exit}\!\setminus\! E^{(i)}_{\exit}$ have their preceding places empty in $m'_3$: $m'_3(r^j)\!=\!0$, $i\!\neq\! j$.
		
		In workflow net $N'$ it holds that $m'_3(r^i)=1$ and $m'_3(p^*)=1$
		($p^*$ is a choice place for the transitions in a free-choice relation within $N$, see Figure~\ref{fig:theorems3_b}) and hence a step: $m'_3\stackrel{a}{\rightarrow}m'_4$, where $a\in E^{(i)}_{\exit}$ can be performed. After $a$ is fired the place $r^i$ is emptied. A corresponding firing step in $N$: $m_3\stackrel{a}{\rightarrow}m_4$ can be taken, because $m_3(p^*)=1$, and all the transitions labelled by events from $E_{\exit}$ are enabled in $m_3$. These steps lead models to the same markings: $\forall p\in P: m_4(p)=m'_4(p)$ from which firing the same transitions the final marking $[o]$ can be reached. If the rest sequence of firings contains events from $E_{\ent}$ and $E_{\exit}$, we repeat the same reasoning.
		
		Thus, we have shown that all the transitions within $N'$ can be fired. Due to the soundness of $N$, since all the firing sequences of $N'$  correspond to firing sequences of $N$, and the number of tokens in each place from $P$ in corresponding markings of $N'$ and $N$ coincide, the final marking can be reached from any reachable marking of $N'$ and there are no reachable markings in $N'$ with tokens in the final place $o$ and some other places.
	\end{proof}
\end{theorem}

\subsection{Using high-level constructs to model discovered non-local constraints}

\begin{figure}[h!]
	\centering
	\includegraphics[scale=0.6]{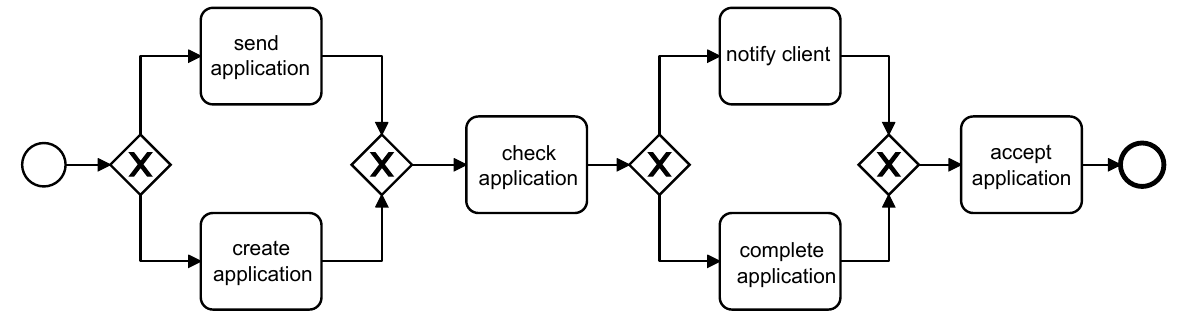}\vspace*{-1mm}
	\caption{A BPMN model that corresponds to the workflow net in Figure~\ref{fig:example}.}
	\label{fig:bpmn1}
\end{figure}

In this subsection, we demonstrate how the discovered process models with non-local constraints can be presented using high-level modelling languages, such as BPMN (Business Process Model and Notation)~\cite{omg2013bpmn202}. Free-choice workflow nets can be modelled by a core set of
process modelling elements that includes start and end events, tasks, parallel and choice gateways, and sequence flows. The equivalence of free-choice workflow nets and process models based on the core set of elements is studied in~\cite{FAVRE2015197,10.1007/3-540-47961-9_37}.
Most process modelling languages, such as BPMN, support these core elements.
A BPMN model corresponding to the discovered free-choice workflow net (shown in Figure~\ref{fig:example}) is presented in Figure~\ref{fig:bpmn1}.

\begin{figure}[!b]
	\centering
	\includegraphics[scale=0.6]{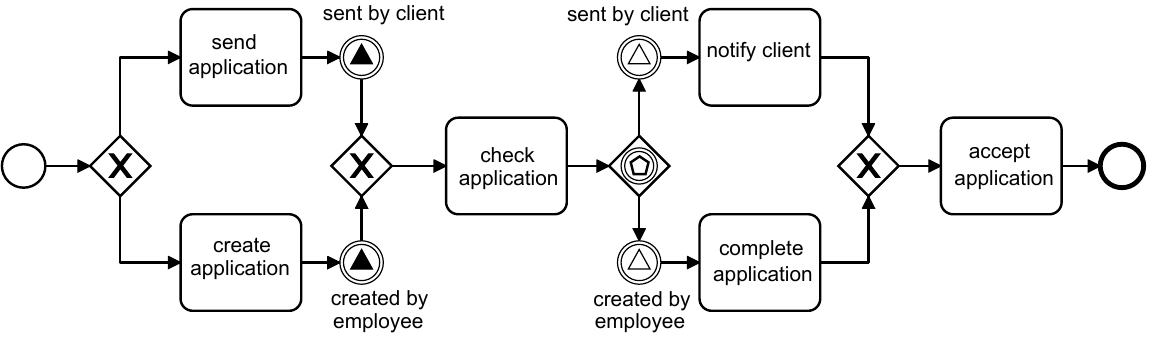}
	\caption{A BPMN model with signals corresponding to the workflow net presented in Figure~\ref{fig:example_enhanced}.}
	\label{fig:bpmn2}\vspace{-1mm}
\end{figure}

If a workflow net is not free-choice, it cannot be presented using core elements only~\cite{FAVRE2015197}. However, \newtext{the} BPMN language offers additional high-level modelling constructs which can be used to model non-free-choice constraints. Figure~\ref{fig:bpmn2} demonstrates a BPMN model that corresponds to a non-free-choice net (in Figure~\ref{fig:example_enhanced}) constructed by Algorithm~\ref{alg:distance}.

\medskip
In addition to core modelling elements, \emph{signal events} and an \emph{event-based gateway} can be used. {\color{black}{Assuming throw signal events are buffered within the scope of the corresponding process instance (this depends on the chosen BPMN engine),}} the signal events can  capture the discovered non-local dependencies. For instance, after \newtext{the} $\mathit{send\,application}$ task is performed, a signal $\mathit{sent\,by\,client}$ is thrown.
After that, an \emph{event-based gateway} is used to select a branch depending on which of \newtext{the catching signal events that immediately follow the gateway} is fired. For example, if the type of the caught event is signal and its value is $\mathit{sent\,by\,client}$, then task $\mathit{notify\,client}$ is  performed.

\begin{figure}[h!]
	\centering
	\includegraphics[scale=0.6]{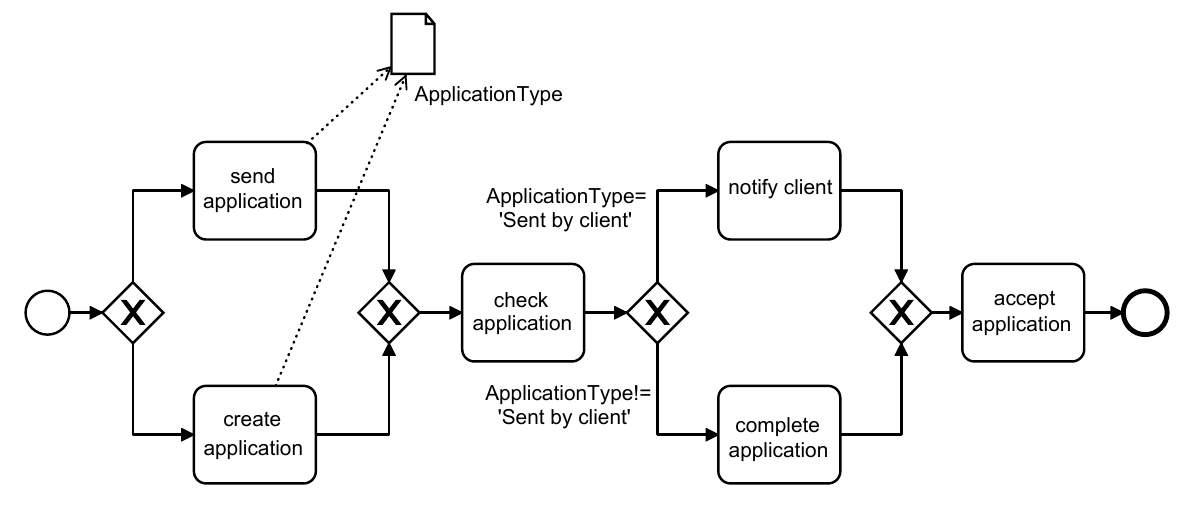}
	\caption{A BPMN model with a data object corresponding to the workflow net presented in Figure~\ref{fig:example_enhanced}.}
	\label{fig:bpmn3}
	\vspace{-5mm}
\end{figure}

\medskip
{\color{black}Besides that, the long-distance dependencies can be modelled in BPMN using data objects and conditional sequence flows~\cite{10.1007/978-3-642-40176-3_14, d17a1e4189e4424bb5d44264f5e24a7c}. Figure~\ref{fig:bpmn3} shows the same process where $\mathit{send\,application}$ and $\mathit{create\,application}$  tasks define the value of \newtext{the} $\mathit{ApplicationType}$ data object, and then, depending on its value one of the process branches is activated.}

\section{Case study}
\label{sec:eval}

In this section, we demonstrate the results of applying our approach to synthetic and real-life event logs. The approach is implemented as an Apromore~\cite{la2011apromore} plugin called \emph{``Add long-distance relations''} and is available as part of \newtext{the} Apromore Community Edition.~\footnote{\url{https://github.com/apromore/ApromoreCE_ExternalPlugins}} All the results were obtained in real-time using Intel(R) Core(TM) i7-8550U CPU @1.80 GHz with 16 GB RAM.

\subsection{Synthetic event logs}

\begin{figure}[h!]
	\vspace{-7pt}
	\centering
	\includegraphics[scale=0.56]{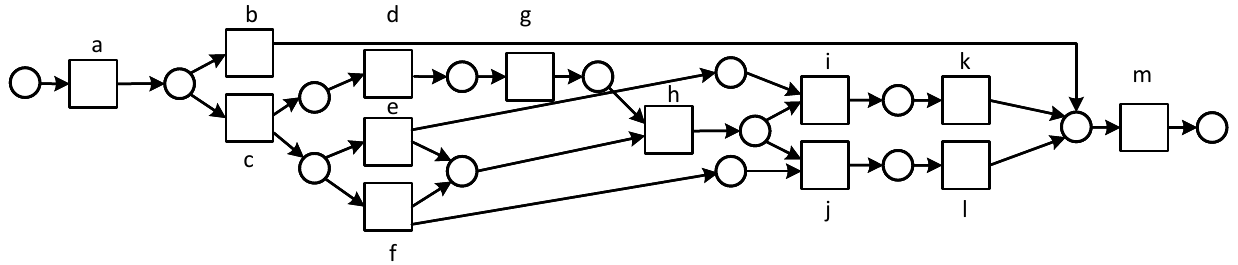}
	\caption{A workflow net used for the synthesis of an event log.}
	\label{fig:synthetic}\vspace{-3mm}
\end{figure}

To assess the ability of our approach to automatically repair process models we have built a set of workflow nets with non-local  dependencies. An example of one of these workflow nets is presented in Figure~\ref{fig:synthetic}.

We simulated each of the workflow nets and generated event logs containing accepted traces. After that, from each event log $L$ we discovered a free-choice workflow net $N$ using Split miner. Then, our approach was applied to $N$ and $L$ producing an enhanced workflow net $N'$ with additional constraints. To compare behaviours of $N$ and $N'$ workflow nets, conformance checking techniques~\cite{DBLP:journals/corr/abs-1812-07334} assessing fitness (the share of the log behaviour accepted by a model) and precision (the share of the model behaviour captured by the log) were applied. In all the cases, both models $N$ and $N'$ accept all the traces from $L$  showing maximum fitness values of 1.0 (according to Theorem~\ref{theorem:fitness}, if $N$ accepts
a~trace,~then $N'$ also accepts this trace). Precision values as well as the structural characteristic of the
workflow nets are presented in Table~\ref{tab:synth_experiment}. These results demonstrate that our approach is able to
 automatically reveal hidden non-local constraints discovering precise workflow nets when applied to synthetic event logs.

  \begin{table}[h!]
		\vspace{-4pt}
          \caption{Structural (number of transitions and number of places) and behavioural characteristics (precision) of free-choice ($N$)
          and enhanced ($N'$) workflow nets. \label{tab:synth_experiment}}
		\centering
		\vspace{-5pt}
		\scalebox{0.9}{
   \begin{tabular}{|| c | c | c | c | c ||}
			\hline
			Event log & \#Transitions / & \#Transitions / & Precision  & Precision \\ [0.5ex]
			& \#Places in $N$  & \#Places in $N'$ & $(N, L)$  & $(N', L)$ \\ [0.5ex]
			\hline\hline
			1 & 18 / 14 & 18 / 18 & 0.972 & 1.0  \\
			\hline
			2 & 13 / 12 & 13 / 14  & 0.945 & 1.0 \\
			\hline
			3 & 10 / 9 & 10 / 11 &   0.899 & 1.0 \\
			\hline
			4 & 12 / 13 & 12 / 15 & 0.911 & 1.0  \\
			\hline
			5 & 6 / 4 & 6 / 6 & 0.841 & 1.0 \\
			\hline
			\hline
		\end{tabular} }\vspace*{-2mm}
	\end{table}

\medskip
At the same time, while other approaches for the discovery of non-free-choice workflow nets, such as $\alpha$++ Miner~\cite{10.1007/s10618-007-0065-y} and the original Petri net synthesis technique~\cite{van2010process} can also synthesize precise workflow nets \newtext{from} this set of simple event logs, they often either produce unsound workflow nets with dead transitions (in \newtext{the} case of $\alpha$++ Miner), or fail to construct a model in \newtext{reasonable} time (in case of the original synthesis approach) when applied to real-world event logs. In the next subsection, we apply our approach to a real-world event log showing that our approach can discover a more precise and sound process model in a real-world setting.

{\color{black}
\subsection{Real-life event logs}

The proposed approach was applied to Business Process Intelligence Challenge (BPIC) real-world event logs, including: (1) the event logs of a university travel expense claims system that processes \emph{Domestic Declarations} (DD-BPIC20)~\cite{van_dongen_2020_dd}, \emph{Prepaid Travel Costs} (PTC-BPIC20)~\cite{van_dongen_2020_ptc}, and \emph{Request For Payment} (RP-BPIC20)~\cite{van_dongen_2020_rp} documents; (2) the \emph{Hospital Billing} (HB-BPIC17)~\cite{mannhardt_2017} event log obtained from financial modules of a hospital information system; (3) the event log of a \emph{Road Traffic Fine Management} (RTF-BPIC15)~\cite{de_leoni_mannhardt_2015} system; (4) the \emph{Receipt phase event log of a building permit application process} (RPBP-BPIC14)~\cite{buijs_2014}; (5)~the \emph{Detailed Incident Activity} (DIA-BPIC14)~\cite{van_dongen_2014} event log of an ITIL (Information Technology Infrastructure Library) process that aligns IT services and banking procedures.

\medskip
To reduce noise and apply the proposed technique to the most frequent behaviour, we filtered each event log, keeping only 20 of its most frequent traces.\footnote{In contrast to the definition of \newtext{event logs} given in Section~\ref{sec:prelim}, in real-life event logs, traces can appear multiple times.}  Then, for each filtered event log $L$, the Inductive mining algorithm~\cite{sander-tree-disc-PN2014} guaranteeing perfect fitness (the language of the discovered model contains all traces from the event log) was applied and a corresponding free-choice workflow net $N$ was discovered. For each pair $(N,L)$, we applied our repair algorithm (Algorithm~\ref{alg:distance}) and constructed a workflow net $N'$ enhanced with additional places. Table~\ref{tab:real_experiments} presents the characteristics of each log~$L$, such as the overall number of occurrences of events and traces, the number of events, the structural characteristics of workflow nets, such as the size of $N$ (the number of places and transitions) and the number of places added to $N'$, the behavioural characteristics (precision) of $N$ and $N'$ with respect to~$L$, and the overall computation time in milliseconds.

These results demonstrate that although we add extra places, increasing the size of the workflow net, we also improve its precision by imposing additional behavioural constraints. To calculate precision we apply the entropy-based conformance checking metric~\cite{DBLP:journals/corr/abs-1812-07334} that is monotonic, i.e., the lower the share of model traces that are not present in the log, the higher the precision. At the same time, we preserve model fitness. According to Theorem~\ref{theorem:fitness},  since the fitness of the initial workflow net is 1.0 (the workflow net accepts all the event log traces), the fitness of the corresponding enhanced workflow net is also~1.0. All the enhanced workflow nets were constructed in less than a second for each of~the~event~logs.

	\begin{table}[h!]
		\footnotesize
		\vspace{-5pt}
		\centering
\caption{Event log characteristics (number of occurrences of events and traces, number of events), size (the number of nodes) of free-choice workflow net $N$, number of new places in enhanced workflow net $N'$, behavioural characteristics (precision) of $N$ and $N'$ in respect to $L$, and calculation time in milliseconds.	\label{tab:real_experiments}}\vspace{-5pt}
		\scalebox{0.98}{
\begin{tabular}{|| c | c | c | c | c | c | c | c | c ||}
			\hline
			Event & \#Events' & \#Traces' & \#Unique  & Size & \#New & Prec. &  Prec. & Time \\  [0.5ex]
			log & occur. & occur. & Events & $N$  & Places & $(N,L)$ & $(N',L)$ & (ms) \\  [0.5ex]
			\hline\hline
			DD-BPIC20 & 54,863 & 10,313 & 14 & 72  & 14 & 0.270 & 0.487 & 133\\
			\hline
			PTC-BPIC20 & 14,448 & 1,727  & 18 & 67 & 26 & 0.233 & 0.397 & 252\\
			\hline
			RP-BPIC20 & 35,342 & 6,694 & 16 & 76 & 14 & 0.317 & 0.529 & 213\\
			\hline
			HB-BPIC17 & 401,718 & 94,056 & 13 & 69 & 3 & 0.310 & 0.393 & 142\\
			\hline
			RTF-BPIC15 & 552,379 & 149,145 & 11 & 44 & 4 & 0.340 & 0.434 & 57\\
			\hline
			RPBP-BPIC14 & 7,526 & 1,328 & 15 & 47 & 6 & 0.463 & 0.546 & 120\\
			\hline
			DIA-BPIC14 & 43,828 & 10,560 & 10 & 31 & 1 & 0.851 & 0.873 & 23\\
			\hline
			\hline
		\end{tabular} }		
	\end{table}

The workflow net $N$ discovered from the real-world event log $L$ of DIA-BPIC14 is presented in Figure~\ref{fig:real_model}. The repair algorithm applied to the model $N$ and event log $L$ constructs $N'$ by discovering a \emph{false free-choice relation} for the activities \emph{Quality Indicator Fixed}, \emph{Resolved} and \emph{Update} and adds a place (region) $r$ that also specifies \emph{Caused by CI} (\emph{Caused by Configuration Item)} as an exit event. This new place restricts the model behaviour in such a way that \newtext{the} \emph{Caused by CI} event can occur only if \emph{Resolved} and \emph{Update} events did not occur. According to the synthesis algorithm (Section~\ref{sec:state}), in addition to the initial $m_0$ and finial $m_f$ markings of $N$ with one token in places $p_0$ and $p_{12}$ respectively, $N'$ assumes a final marking $m'_f$, such that $m'_f(p_{12})=m'_f(r)=1$. In contrast to $N$, $N'$ is not sound because a marking with tokens in places $p_{12}$ and $r$ is reachable. However, according to Theorem~\ref{theorem:fitness}, $N'$ can replay all the traces from $L$; also, $N'$ is more precise than $N$, in respect to $L$, i.e., precision for $N$ is 0.851, while precision for $N'$ is 0.873 (Table~\ref{tab:real_experiments}).

\begin{figure}[h!]
	\vspace{-12pt}
	\centering
	\includegraphics[scale=0.55]{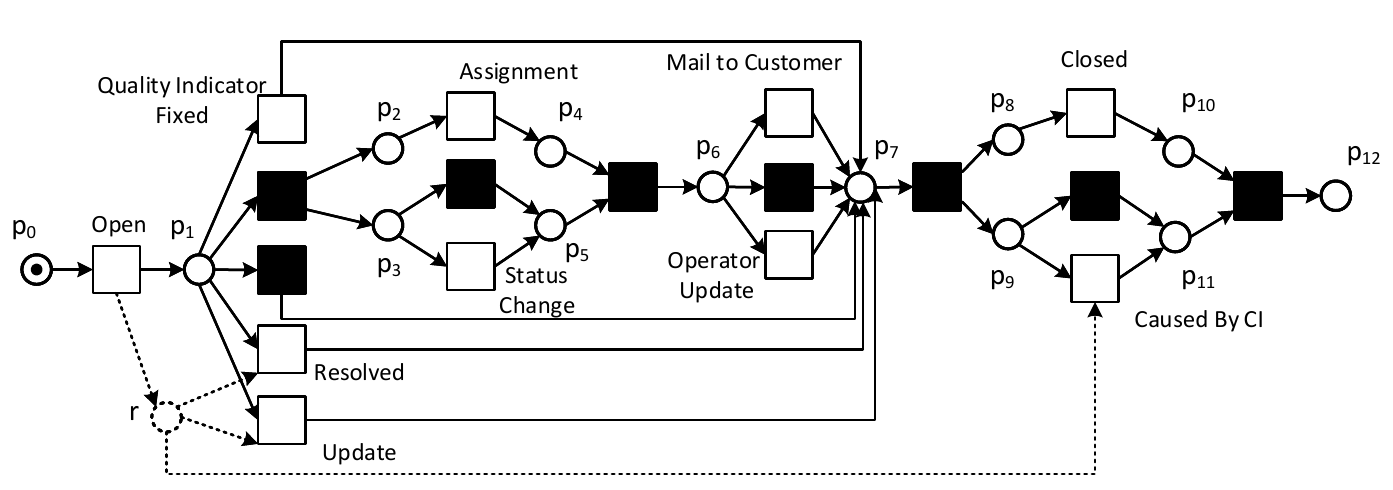}\vspace{1mm}
	\caption{The workflow net $N$ discovered from DIA-BPIC14 event log using the Inductive mining algorithm that guarantees perfect fitness and an enhanced model $N'$ with an additional place $r$ discovered by the repair algorithm. The initial marking $m_0$ of $N'$ is the marking with only one token in place $p_0$, i.e., $m_0(p_0)=1$, for two final markings $m_f$ and $m'_f$ of $N'$, it holds that $m_f(p_{12})=m'_f(p_{12})=1$ and, additionally $m'_f(r)=1$.}
	\label{fig:real_model}
	\vspace{-10pt}
\end{figure}

\newtext{To analyse the calculation times, we compared the proposed repair technique to the region-based approaches that discover process models from event logs without constructing intermediate process representations (Table~\ref{tab:real_experiments2}). To that end, we applied the region synthesis algorithm~\cite{CarmonaCK10} to the transition systems constructed from the real-world event logs described in Table~\ref{tab:real_experiments}. The discovered Petri nets $\overline{N}$ were obtained in a reasonable time that was higher than the time of applying the proposed repair technique, because not only the regions that solve the outlined $\ESSP$ problems were discovered, but also other places. The Petri net models $\overline{N}$ are more compact than the original workflow nets $N$ and $N'$, because they do not contain silent transitions. However, the structure of these Petri nets is not presented by solid control flow graphs and they usually contain unconnected transitions. For instance, the Petri net discovered from the event log DIA-BPIC14 contains five transitions  \emph{Quality Indicator Fixed}, \emph{Assignment},  \emph{Status Change}, \emph{Mail to Customer} and \emph{Operator Update} that do not have any incoming or outgoing flows. To discover Petri nets that are represented by a solid control flow, the synthesis algorithm~\cite{deriving} that incorporates the label splitting procedure can be applied. We synthesised workflow nets $\doverline{N}$ with duplicate transitions (transitions with the identical labels) from the real-world event logs (Table~\ref{tab:real_experiments}) using the ProM (Process Mining Framework)~\footnote{\url{https://www.promtools.org/.}} plugin \emph{Convert to Petri net using Regions}~\cite{van2010process}. As follows from Table~\ref{tab:real_experiments2}, this technique is time consuming and some of the event logs could not be analysed in a reasonable amount of time. The discovered models $\doverline{N}$ are not block-structured and contain duplicate transitions, however, they are comparable in size to the repaired models $N'$ and clearly represent the control flow structure. Comparing to the other region-based techniques, our repair approach achieves two goals: it constructs structured process models (with additional constraints) in a feasible time.}

	\begin{table}[h!]
			\vspace{-7pt}
        \caption{\newtext{The size (the number of nodes) and the discovery time (in milliseconds) for Petri nets $\overline{N}$ and workflow nets with duplicated labels  $\protect\doverline{N}$, discovered using region-based techniques from the event logs described in Table~\ref{tab:real_experiments}.}}
		\label{tab:real_experiments2}\vspace*{-2.5mm}
		\centering
		\scalebox{0.88}{
         	\begin{tabular}{|| c | c | c  | c | c ||}
			\hline
			Event  & Size & Disc. time & Size  & Disc. time \\  [0.5ex]
			log & $\overline{N}$  & $\overline{N}$ (ms) & $\doverline{N}$ & $\doverline{N}$ (ms) \\  [0.5ex]
			\hline\hline
			DD-BPIC20  & 45  & 669 & -- & --\\
			\hline
			PTC-BPIC20 & 63 & 2,761 & -- & --\\
			\hline
			RP-BPIC20 & 51 & 1,056 & -- & --\\
			\hline
			HB-BPIC17 & 37 & 482 & 51 & 630,308\\
			\hline
			RTF-BPIC15 & 32 & 487 & 43 & 13,258\\
			\hline
			RPBP-BPIC14 & 30 & 417 & 62 & 259,642\\
			\hline
			DIA-BPIC14 & 19 & 235 & 39 & 2,462\\
			\hline
			\hline
		\end{tabular} }	\vspace*{-3mm}
	\end{table}

}

\section{Conclusion and future work}
\label{sec:concl}

This paper presents an automated repair approach for obtaining precise process models under the presence of non-local dependencies. The approach identifies opportunities for improving the process model by analysing the process behaviour recorded in the input event log. It then uses goal-oriented region-based synthesis to discover new Petri net fragments that introduce non-local dependencies.

The theoretical contributions of this paper have been implemented as an open-source plugin of the Apromore process mining platform. This implementation has then been used to provide experimental results.  {\color{black}Based on the experiments conducted, the proposed approach shows good performance.}

We foresee different research directions arising from this work. First, implementing the proposed approach for alternative region techniques like language-based~\cite{10.1007/978-3-540-68746-7_24,BergenthumDLM08} or geometric~\cite{BestDS17,SchlachterW18} is an interesting avenue to explore. Second, evaluating the impact that well-known problems with event logs, like {\em noise} or {\em incompleteness}, may have on the approach, and proposing possible ways to overcome these problems should be explored.  {\color{black}Finally, we plan to investigate and classify behavioural characteristics of unsound models discovered  by our repair approach.}


\begin{thebibliography}{10}
\providecommand{\url}[1]{\texttt{#1}}
\providecommand{\urlprefix}{URL }
\expandafter\ifx\csname urlstyle\endcsname\relax
  \providecommand{\doi}[1]{doi:\discretionary{}{}{}#1}\else
  \providecommand{\doi}{doi:\discretionary{}{}{}\begingroup
  \urlstyle{rm}\Url}\fi
\providecommand{\eprint}[2][]{\url{#2}}

\bibitem{Aalst16}
van~der Aalst W.
\newblock Process mining: data science in action.
\newblock Springer, 2016.
\newblock ISBN 978-3-662-49850-7.

\bibitem{CarmonaDSW18}
Carmona J, van Dongen B, Solti A, Weidlich M.
\newblock Conformance checking - relating processes and models.
\newblock Springer, 2018.
\newblock ISBN:978-3-319-99413-0.

\bibitem{omg2013bpmn202}
OMG.
\newblock {Business Process Model and Notation (BPMN), Version 2.0.2}, 2013.
\newblock \urlprefix\url{http://www.omg.org/spec/BPMN/2.0.2}.

\bibitem{sander-tree-disc-PN2014}
Leemans S, Fahland D, van~der Aalst W.
\newblock Discovering block-structured process models from incomplete event  logs.
\newblock In: ATPN'2014, volume 8489 of \emph{LNCS}, pp. 91--110. Springer,
  2014.   doi:10.1007/978-3-319-07734-5\_6.

\bibitem{10.1007/s10115-018-1214-x}
Augusto A, Conforti R, Dumas M, La~Rosa M, Polyvyanyy A.
\newblock Split Miner: Automated discovery of accurate and simple business
  process models from event logs.
\newblock \emph{Knowl. Inf. Syst.}, 2019.
\newblock \textbf{59}(2):251–284.  doi:10.1007/s10115-018-1214-x.

\bibitem{free-choice}
Desel J, Esparza J.
\newblock Free choice Petri nets.
\newblock Cambridge University Press, USA, 1995.
\newblock ISBN: 0521465192.

\bibitem{sosym}
Kalenkova A, van~der Aalst W, Lomazova I, Rubin V.
\newblock {Process mining using BPMN: Relating event logs and process models
  process mining using BPMN.}
\newblock \emph{Software and Systems Modeling}, 2017.
\newblock \textbf{16}:1019--1048.   doi:10.1007/s10270-015-0502-0.

\bibitem{10.1007/s10618-007-0065-y}
Wen L, Aalst W, Wang J, Sun J.
\newblock Mining process models with non-free-choice constructs.
\newblock \emph{Data Min. Knowl. Discov.}, 2007.
\newblock \textbf{15}(2):145–180.  doi:10.1007/s10618-007-0065-y.

\bibitem{10.1007/978-3-540-85758-7_26}
Carmona J, Cortadella J, Kishinevsky M.
\newblock A region-based algorithm for discovering Petri nets from event logs.
\newblock In: Business Process Management. Springer Berlin Heidelberg, Berlin,
  Heidelberg.
\newblock 2008 pp. 358--373.  ISBN:978-3-540-85758-7.

\bibitem{van2010process}
van~der Aalst W, Rubin V, Verbeek H, van Dongen B, Kindler E, G{\"u}nther C.
\newblock Process mining: a two-step approach to balance between underfitting
  and overfitting.
\newblock \emph{Software \& Systems Modeling}, 2010.
\newblock \textbf{9}(1):87.  doi:10.1007/s10270-008-0106-z.

\bibitem{10.1007/978-3-642-13675-7_14}
Sol\'{e} M, Carmona J.
\newblock Process mining from a basis of state regions.
\newblock In: Proceedings of the 31st International Conference on Applications
  and Theory of Petri Nets, PETRI NETS’10. Springer-Verlag, Berlin,
  Heidelberg.
\newblock  2010 p. 226–245. ISBN:3642136745.

\bibitem{10.1007/978-3-540-68746-7_24}
van~der Werf J, van Dongen B, Hurkens C, Serebrenik A.
\newblock Process discovery using integer linear programming.
\newblock In: Applications and Theory of Petri Nets. Springer Berlin
  Heidelberg, Berlin, Heidelberg. 2008 pp. 368--387.
\newblock ISBN:978-3-540-68746-7.

\bibitem{prime}
{Bergenthum} R.
\newblock Prime miner - process discovery using prime event structures.
\newblock In: 2019 International Conference on Process Mining (ICPM). 2019 pp.
  41--48.  doi:10.1109/ICPM.2019.00017.

\bibitem{DBLP:journals/computing/ZelstDAV18}
van Zelst S, van Dongen B, van~der Aalst W, Verbeek H.
\newblock Discovering workflow nets using integer linear programming.
\newblock \emph{Computing}, 2018.
\newblock \textbf{100}(5):529--556.  doi:10.1007/s00607-017-0582-5.

\bibitem{est}
Mannel L, van~der Aalst W.
\newblock Finding uniwired Petri nets using eST-Miner.
\newblock In: Business Process Management Workshops. Springer International
  Publishing, Cham. 2019 pp. 224--237.
\newblock ISBN:978-3-030-37453-2.

\bibitem{DBLP:conf/apn/MannelA19}
Mannel L, van~der Aalst W.
\newblock Finding complex process-structures by exploiting the token-game.
\newblock In: 40th International Conference, {PETRI} {NETS}'19, Proceedings,
  volume 11522 of \emph{Lecture Notes in Computer Science}. Springer, 2019 pp.
  258--278.  doi:10.1007/978-3-030-21571-2\_15.

\bibitem{BadouelBD15}
Badouel E, Bernardinello L, Darondeau P.
\newblock Petri net synthesis.
\newblock Texts in Theoretical Computer Science. An {EATCS} Series. Springer,
  2015.
\newblock ISBN:978-3-662-47966-7.

\bibitem{BergenthumDLM08}
Bergenthum R, Desel J, Lorenz R, Mauser S.
\newblock Synthesis of Petri nets from finite partial languages.
\newblock \emph{Fundam. Inform.}, 2008.
\newblock \textbf{88}(4):437--468.

\bibitem{BestDS17}
Best E, Devillers R, Schlachter U.
\newblock A graph-theoretical characterisation of state separation.
\newblock In: {SOFSEM} - 43rd International Conference on Current Trends in
  Theory and Practice of Computer Science, Proceedings. 2017 pp. 163--175.
  doi:10.1007/978-3-319-51963-0\_13.

\bibitem{SchlachterW18}
Schlachter U, Wimmel H.
\newblock A geometric characterisation of event/state separation.
\newblock In: Application and Theory of Petri Nets and Concurrency - 39th
  International Conference, {PETRI} {NETS}'18, Proceedings. 2018 pp. 99--116.

\bibitem{10.1145/2980764}
Polyvyanyy A, van~der Aalst W, ter Hofstede A, Wynn M.
\newblock Impact-driven process model repair.
\newblock \emph{ACM Trans. Softw. Eng. Methodol.}, 2016.
\newblock \textbf{25}(4). doi:10.1145/2980764.

\bibitem{10.1007/978-3-319-69462-7_5}
Armas-Cervantes A, van Beest N, La~Rosa M, Dumas M, Garc{\'i}a-Ba{\~{n}}uelos  L.
\newblock Interactive and incremental business process model repair.
\newblock In: On the Move to Meaningful Internet Systems. OTM 2017 Conferences.
  Springer International Publishing, Cham.  2017 pp. 53--74.
\newblock ISBN:978-3-319-69462-7.  doi:10.1007/978-3-319-69462-7\_5.

\bibitem{FAHLAND2015220}
Fahland D, van~der Aalst W.
\newblock Model repair — aligning process models to reality.
\newblock \emph{Information Systems}, 2015.
\newblock \textbf{47}:220--243. doi:10.1016/j.is.2013.12.007.

\bibitem{d429471f542a45f09a62131cf704e730}
Mitsyuk A, Lomazova I, Shugurov I, {van der Aalst} W.
\newblock Process model repair by detecting unfitting fragments.
\newblock In: AIST 2017, CEUR Workshop Proceedings. 2017 pp. 301--313.
doi:d429471f542a45f09a62131cf704e730.

\bibitem{la2011apromore}
La~Rosa M, Reijers H, van Der~Aalst W, Dijkman R, Mendling J, Dumas M,
  Garc{\'\i}a-Ba{\~n}uelos L.
\newblock APROMORE: An advanced process model repository.
\newblock \emph{Expert Systems with Applications}, 2011.
\newblock \textbf{38}(6):7029--7040.  doi:10.1016/j.eswa.2010.12.012.

\bibitem{petrinets2020}
Kalenkova A, Carmona J, Polyvyanyy A, {La Rosa} M.
\newblock Automated repair of process models using non-local constraints.
\newblock In: 41st International Conference, {PETRI} {NETS}'20, Proceedings,
  volume 12152 of \emph{Lecture Notes in Computer Science}. Springer, 2020 pp.
  280--300.

\bibitem{707587}
{Cortadella} J, {Kishinevsky} M, {Lavagno} L, {Yakovlev} A.
\newblock Deriving Petri nets from finite transition systems.
\newblock \emph{IEEE Transactions on Computers}, 1998.
\newblock \textbf{47}(8):859--882.  doi:10.1109/12.707587.

\bibitem{CarmonaCK10}
Carmona J, Cortadella J, Kishinevsky M.
\newblock New region-based algorithms for deriving bounded Petri nets.
\newblock \emph{{IEEE} Trans. Computers}, 2010.
\newblock \textbf{59}(3):371--384.  doi:10.1109/TC.2009.131.

\bibitem{hopcroft}
Hopcroft J, Ullman J.
\newblock An n log n algorithm for detecting reducible graphs.
\newblock In: Proe. 6th Annual Princeton Conf. on Inf. Sciences and Systems.
  1972 pp. 119--122.

\bibitem{10.1007/3-540-47961-9_37}
van~der Aalst W, Hirnschall A, Verbeek H.
\newblock An alternative way to analyze workflow graphs.
\newblock In: Advanced Information Systems Engineering. Springer Berlin
  Heidelberg. 2002 pp. 535--552.
\newblock ISBN:978-3-540-47961-1.  doi:10.1007/3-540-47961-9\_37.

\bibitem{DeselR96}
Desel J, Reisig W.
\newblock The synthesis problem of Petri nets.
\newblock \emph{Acta Inf.}, 1996.
\newblock \textbf{33}(4):297--315.  doi:10.1007/s002360050046.

\bibitem{deriving}
{Cortadella} J, {Kishinevsky} M, {Lavagno} L, {Yakovlev} A.
\newblock Deriving Petri nets from finite transition systems.
\newblock \emph{IEEE Transactions on Computers}, 1998.
\newblock \textbf{47}(8):859--882.

\bibitem{10.5555/3405}
Reisig W.
\newblock Petri nets: An introduction.
\newblock Springer-Verlag, Berlin, Heidelberg, 1985.
\newblock ISBN: 0387137238.  doi:10.5555/3405.

\bibitem{FAVRE2015197}
Favre C, Fahland D, Völzer H.
\newblock The relationship between workflow graphs and free-choice workflow  nets.
\newblock \emph{Information Systems}, 2015.
\newblock \textbf{47}:197 -- 219.

\bibitem{10.1007/978-3-642-40176-3_14}
Meyer A, Pufahl L, Fahland D, Weske M.
\newblock Modeling and enacting complex data dependencies in business
  processes. 2013 pp. 171--186.
\newblock In: Business Process Management. Springer Berlin Heidelberg.
\newblock ISBN:978-3-642-40176-3.  doi:10.1007/978-3-642-40176-3\_14.

\bibitem{d17a1e4189e4424bb5d44264f5e24a7c}
Kalenkova A, Burattin A, {de Leoni} M, {van der Aalst} W, Sperduti A.
\newblock Discovering high-level BPMN process models from event data.
\newblock \emph{Business Process Management Journal}, 2019.
\newblock \textbf{25}(5):995--1019.  doi:10.1108/BPMJ-02-2018-0051.

\bibitem{DBLP:journals/corr/abs-1812-07334}
Polyvyanyy A, Solti A, Weidlich M, Di~Ciccio C, Mendling J.
\newblock Monotone precision and recall measures for comparing executions and
  specifications of dynamic systems.
\newblock \emph{ACM Trans. Softw. Eng. Methodol.}, 2020.
\newblock \textbf{29}(3):1--41.  doi:10.1145/3387909.

\bibitem{van_dongen_2020_dd}
van Dongen B.
\newblock BPI challenge 2020: Domestic declarations, 2020.
\newblock doi:10.4121/uuid:3f422315- ed9d-4882-891f-e180b5b4feb5.

\bibitem{van_dongen_2020_ptc}
van Dongen B.
\newblock BPI challenge 2020: Prepaid travel costs, 2020.
\newblock doi:10.4121/uuid:5d2fe5e1- f91f-4a3b-ad9b-9e4126870165.

\bibitem{van_dongen_2020_rp}
van Dongen B.
\newblock BPI challenge 2020: Request for payment, 2020.
\newblock doi:10.4121/uuid:895b26fb- 6f25-46eb-9e48-0dca26fcd030.

\bibitem{mannhardt_2017}
Mannhardt F.
\newblock Hospital billing - event log, 2017.
\newblock doi:10.4121/uuid:76c46b83- c930-4798-a1c9-4be94dfeb741.

\bibitem{de_leoni_mannhardt_2015}
de~Leoni M, Mannhardt F.
\newblock Road traffic fine management process, 2015.
\newblock doi:10.4121/uuid: 270fd440-1057-4fb9-89a9-b699b47990f5.

\bibitem{buijs_2014}
Buijs J.
\newblock Receipt phase of an environmental permit application process
  (‘WABO’), CoSeLoG project, 2014.
\newblock doi:10.4121/uuid:a07386a5- 7be3-4367-9535-70bc9e77dbe6.

\bibitem{van_dongen_2014}
van Dongen B.
\newblock BPI challenge 2014: Activity log for incidents, 2014.
\newblock doi:10.4121/uuid: 86977bac-f874-49cf-8337-80f26bf5d2ef.
\end{thebibliography}


\end{document}